\documentclass[sigplan,10pt,nonacm]{acmart}

\setcopyright{none}

\usepackage{booktabs}
\usepackage{graphicx}
\usepackage{amsmath,amsfonts}
\usepackage{xcolor}
\usepackage{balance}
\usepackage{placeins}
\graphicspath{{figures/}}

\newcommand{\sys}{\textsc{SuffixReplay}}

\begin{document}

\title[Just Let Linear States Forget the Distant Past]{Just Let Linear States Forget the Distant Past: Prefix Caching via SuffixReplay for Hybrid LLMs}

\author{Yirui Liu}
\authornote{Equal contribution.}
\authornote{Corresponding authors.}
\email{yiruiliu926@gmail.com}
\affiliation{\institution{Institute of Artificial Intelligence, China Telecom (TeleAI)}\country{China}}

\author{Ruoling Qi}
\authornotemark[1]
\email{qiruoling760@sjtu.edu.cn}
\affiliation{\institution{Shanghai Jiao Tong University}\country{}}
\affiliation{\institution{Institute of Artificial Intelligence, China Telecom (TeleAI)}\country{China}}

\author{Xuaner Wu}
\authornotemark[1]
\email{xuanerwuu@gmail.com}
\affiliation{\institution{Institute of Artificial Intelligence, China Telecom (TeleAI)}\country{China}}

\author{Yuxin Jin}
\email{yuxinkim1007@gmail.com}
\affiliation{\institution{Institute of Artificial Intelligence, China Telecom (TeleAI)}\country{China}}

\author{Jian Chen}
\email{cjcobalt@icloud.com}
\affiliation{\institution{Individual Researcher}\country{United States}}

\author{Penghang Liu}
\email{penghangliu5365@gmail.com}
\affiliation{\institution{Individual Researcher}\country{United States}}

\author{Yafei Huang}
\email{huangyf24@mails.tsinghua.edu.cn}
\affiliation{\institution{Tsinghua University}\country{}}
\affiliation{\institution{Institute of Artificial Intelligence, China Telecom (TeleAI)}\country{China}}

\author{Jiawei Shao}
\email{shaojw2@chinatelecom.cn}
\affiliation{\institution{Institute of Artificial Intelligence, China Telecom (TeleAI)}\country{China}}

\author{Xuelong Li}
\authornotemark[2]
\email{xuelong_li@ieee.org}
\affiliation{\institution{Institute of Artificial Intelligence, China Telecom (TeleAI)}\country{China}}

\renewcommand{\shortauthors}{Liu, Qi, Wu, et al.}

\begin{abstract}
Hybrid LLMs interleave full-attention layers with linear-attention layers to reduce long-context inference cost, but this structure complicates prefix caching. Full-attention KV caches are token-addressable, whereas linear-attention layers maintain recurrent states that cannot be rolled back to arbitrary prefix boundaries. Existing systems materialize recurrent-state checkpoints, restricting prefix reuse to checkpoint-aligned positions.

We present \sys{}, to our knowledge the first prefix caching system that enables hybrid LLMs to reuse cached prefixes at every cache-supported page boundary without materializing recurrent-state checkpoints. Our key insight, reflected in the title, is to \emph{just let linear states forget the distant past}. Modern linear-attention mechanisms use recurrent decay and gating to attenuate the influence of sufficiently old inputs. Therefore, instead of storing a recurrent-state checkpoint at every possible prefix boundary, \sys{} approximates the state at a matched boundary by replaying only a recent suffix of the layer's input hidden states, which we retain as anchors. At the algorithmic level, \sys{} combines layer-wise and token-wise anchor sparsity with a bounded replay budget to jointly control storage, computation, and quality. At the system level, it uses an independently managed anchor sidecar and a pipelined replay path to overlap anchor movement and state reconstruction with the native serving pipeline. We evaluate \sys{} on three hybrid LLMs: OLMo-Hybrid-7B, Qwen3.5-4B, and Qwen3.6-27B-FP8. Across these models, \sys{} retains 91.4--100\% of full-prefill quality on average across LongBench and RULER, while using only 0.36--0.51$\times$ the amortized per-token storage of SGLang's default 8192-token checkpoint cache. Integrated into SGLang, \sys{} reduces median TTFT by 15--70\% on branching workloads, sustains 2.3--4.3$\times$ SGLang's throughput when the working set exceeds HBM, and matches SGLang on high-hit continuation traffic.
\end{abstract}

\maketitle

\section{Introduction}
\label{sec:intro}

LLM applications have evolved from single-turn instruction-following assistants~\cite{instructgpt} to richer workloads such as multi-turn dialogue~\cite{lmsyschat}, retrieval-augmented generation~\cite{rag}, tool-using agents~\cite{react}, and multi-agent workflows~\cite{autogen,metagpt}. These workloads bring longer contexts, repeated invocations, and higher concurrency, making serving efficiency an increasingly important practical concern. To address these pressures, model designers and system designers have pursued complementary solutions in their respective domains. At the model level, hybrid architectures~\cite{jamba,samba,nemotronh,minimax01,qwen35,qwen36,glm53flash,olmohybrid} combine full-attention (FA) layers with linear-attention layers, reducing the amount of quadratic FA computation required for long-context inference. At the serving level, prefix caching~\cite{sglang,promptcache,cachedattention,cacheblend} reuses shared prefixes across requests to avoid redundant prefill.

\begin{figure}[t]
  \centering
  \includegraphics[width=\linewidth]{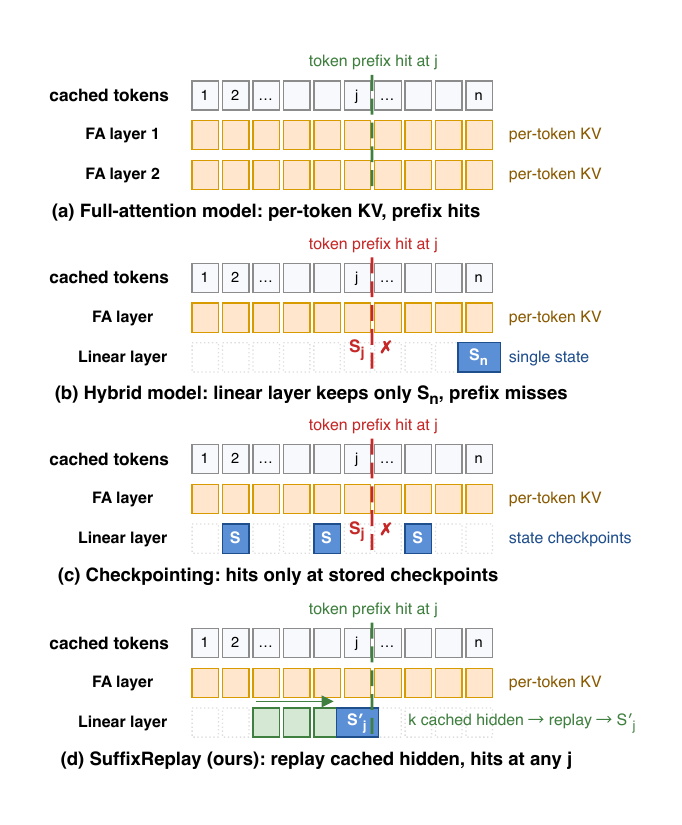}
  \vspace{-0.3in}
  \caption{Prefix caching for FA-only and hybrid models. }
  \label{fig:intro-hit}
  \vspace{-0.1in}
\end{figure}

Hybrid models and prefix caching are two complementary techniques. A natural question is whether they can be co-deployed to achieve greater overall serving efficiency. However, the standard prefix-caching abstraction is designed primarily for FA-only models and does not directly extend to hybrid architectures. In an FA model, the reusable state consists of token-indexed key-value (KV) caches, so the KV cache of a processed prefix can be retrieved and reused at the corresponding prefix boundary (Figure \ref{fig:intro-hit}. a). In contrast, linear-attention layers summarize the processed prefix into recurrent states through in-place updates. Once such a state has advanced to a later position, it cannot be directly rolled back to represent an arbitrary earlier prefix (Figure \ref{fig:intro-hit}. b). 

Existing approaches, including Marconi~\cite{marconi}, Sparse Prefix Caching~\cite{sparseprefix}, and SGLang~\cite{sglang}, reconcile this mismatch by materializing linear-attention state checkpoints at selected positions (Figure~\ref{fig:intro-hit}c). These approaches make hybrid prefix reuse possible, but fundamentally discretize prefix reuse: a shared token prefix is reusable only when its endpoint aligns with a materialized checkpoint. Increasing checkpoint density can mitigate this limitation. If a checkpoint were stored at every token, hybrid prefix caching would recover the token-level reuse granularity of FA-only models. However, this approach is prohibitively expensive. For example, in Qwen3.5-4B, one linear-attention state checkpoint is approximately 49\,MiB, whereas the KV cache for one token is only about 32\,KiB, making the checkpoint roughly 1,500$\times$ larger than a single-token KV entry. Dense checkpoints also increase the amount of state that must be written, transferred, indexed, and managed. The resulting challenge is how to make hybrid prefix caching finer-grained without incurring the substantial storage and management cost of dense recurrent-state checkpoints.

\paragraph{Key Insight.} Our goal is to make prefix reuse in hybrid LLMs as fine-grained as prefix caching in FA-only models: a matched prefix should be reusable at any cache-hit position supported by the cache. Rather than densifying linear-attention state checkpoints, we avoid storing these checkpoints altogether. Instead, we retain the inputs to the recurrent updates and reconstruct the corresponding states on demand at the matched prefix boundary (Figure~\ref{fig:intro-hit}d). Our key insight is twofold, summarized by the paper's title: \emph{just let linear states forget the distant past}.

First, given a sequence of input hidden states $h_1,\ldots,h_n$ to a linear-attention layer, the state corresponding to any prefix position $j\in[1,n]$ is a deterministic function of the inputs consumed by that prefix:
\begin{equation}
  S_j = f(h_1,\ldots,h_j).
\end{equation}
Thus, although $S_n$ cannot be rolled back to $S_j$ algebraically, the state at any prefix position can be reconstructed by replaying the corresponding inputs. However, directly replaying the entire prefix requires $\mathcal{O}(j)$ computation, which grows with the cache-hit position and becomes prohibitive for long contexts.

Second, modern linear-attention mechanisms progressively attenuate the contribution of distant inputs through recurrent decay and erase gates. Thus, the contribution of earlier hidden states to $S_j$ diminishes with their distance from position $j$, allowing us to let the reconstructed state forget the distant past and approximate $S_j$ by replaying only a recent suffix:
\begin{equation}
    S_j\approx S'_j = f(h_{j-k+1},\ldots,h_j).
\end{equation}
If a relatively short suffix is sufficient to make $S'_j$ a valid proxy for $S_j$, replay cost depends on the suffix length $k$ rather than on the full prefix length. This observation opens a new design space for fine-grained prefix reuse in hybrid LLMs while leaving the model itself unchanged.

\paragraph{Design Requirements.}
Suffix replay opens a new design space with three coupled costs. Storing the hidden states needed for replay incurs additional storage, replaying the suffix incurs computation, and approximating $S_j$ with $S'_j$ may affect end-to-end generation quality. We therefore design our system to satisfy the following three requirements:

\noindent \textbf{R1: Preserving quality.}
The reconstructed state $S'_j$ should serve as a sufficiently accurate proxy for the full state $S_j$ to preserve end-to-end generation quality. Across LongBench and RULER, we target at least 90\% of the quality achieved by full prefill on average.

\noindent \textbf{R2: Bounding additional storage.}
Caching the inputs needed for replay should not simply replace the storage cost of linear-state checkpoints with an equivalent hidden-state cache. SGLang's default hybrid-cache configuration stores one linear-state checkpoint every 8192 tokens. Thus, if $B_{\mathrm{ckpt}}$ denotes the size of one checkpoint, its amortized additional storage is $B_{\mathrm{ckpt}}/8192$ per token. We set this value as the upper bound for the additional storage of replay inputs:
\begin{equation}
  B_{\mathrm{anchor}} \leq \frac{B_{\mathrm{ckpt}}}{8192}.
\end{equation}
This is a challenging target because it uses a sparse default checkpoint interval. In practice, supporting finer-grained reuse may require a smaller interval, such as 4096 tokens, which increases the native per-token checkpoint overhead and makes this storage target easier to satisfy.

\noindent \textbf{R3: Maintaining serving performance.}
Replay should not offset the benefit of prefix caching. The resulting system should match or outperform SGLang's native prefix-caching path in TTFT, throughput, and capacity under concurrent requests.

\paragraph{Our approach.}
We present \sys{}, to the best of our knowledge, the first prefix-caching system that gives hybrid LLMs the same prefix-reuse granularity as FA-only models. Through the co-design of the suffix-replay algorithm and its serving system, \sys{} reconstructs linear-attention states at every cache-supported page boundary without materializing recurrent-state checkpoints. At the algorithmic level, \sys{} combines suffix replay with layer-wise and token-wise sparse anchors to bound replay computation and additional storage. At the system level, \sys{} introduces an independently managed anchor sidecar and a replay-aware serving path that coordinates anchor writes, anchor fetching, replay, KV movement, and forward execution.

We evaluate \sys{} on three hybrid LLMs: OLMo-Hybrid-7B, Qwen3.5-4B, and Qwen3.6-27B-FP8. With calibrated replay budgets, OLMo-Hybrid-7B retains 95.9\% of full-prefill quality on LongBench QA and 91.4\% on RULER, while Qwen3.5-4B and Qwen3.6-27B-FP8 each exceed 99\% on average on both benchmarks. Across the three models, \sys{} reduces the additional storage to 0.36--0.51$\times$ that of SGLang's native cache at its default 8192-token checkpoint interval. Integrated into SGLang and evaluated on Qwen3.5-4B and Qwen3.6-27B-FP8, \sys{} reduces median TTFT by 15--70\% on workloads where requests branch from shared prefixes compared with SGLang's native hybrid prefix cache, matches SGLang's native cache on high-hit continuation traffic, and sustains 2.3--4.3$\times$ its throughput when the working set exceeds HBM. These results demonstrate that \sys{} meets the three requirements above.

\section{Background and Motivation}
\label{sec:background}

\subsection{Hybrid LLMs and Linear Attention}
\label{sec:bg-hybrid}

Hybrid LLMs combine full-attention (FA) layers with linear-attention layers to reduce the cost of processing long contexts. An FA layer represents a processed prefix with token-indexed key--value (KV) entries, while a linear-attention layer summarizes the prefix in a fixed-size recurrent state. Let $h_t$ denote the hidden vector entering a linear-attention layer at token $t$. The layer maintains a state $S_t$ and updates it once per token:
\begin{equation}
  S_t = \mathcal{A}_t\!\left(S_{t-1}\right) + B_t, \qquad S_0 = 0,
  \label{eq:recurrence}
\end{equation}
where $\mathcal{A}_t$ is a linear transformation of the previous state and $B_t$ is the write contributed by token $t$. Both terms are computed from $h_t$: the keys, values, and gates used by the update are projections of the layer input. Different linear-attention mechanisms use different forms of $\mathcal{A}_t$. For example, Mamba2 uses a per-head decay, while Gated DeltaNet (GDN) combines decay with a rank-one erase operation~\cite{mamba2,gateddeltanet}. Unrolling Eq.~\eqref{eq:recurrence} gives the state as a function of the layer inputs:
\begin{equation}
  S_n = f(h_1,\ldots,h_n) = \sum_{i=1}^{n} \Phi_{n,i}\!\left(B_i\right), \qquad \Phi_{n,i} = \mathcal{A}_n \circ \cdots \circ \mathcal{A}_{i+1}.
  \label{eq:unrolled}
\end{equation}
The state is therefore a summary of the prefix, updated in place as new tokens arrive. Unlike the KV entries of an FA layer, it does not preserve a separately addressable value for every earlier position.

\subsection{Prefix Caching and Reuse Boundaries}
\label{sec:bg-prefix}

Long-context serving workloads repeatedly process shared prefixes in multi-turn conversations, retrieval-augmented generation, and tool-using agents. Prefix caching avoids this redundant prefill work by retaining the representation of a processed prefix and reusing it when a later request shares the same prefix~\cite{sglang,promptcache,cachedattention,cacheblend}. The reuse boundary is determined by program execution: an agent may branch after a tool call, a program may create several continuations from an intermediate response, and a chat system may regenerate, edit, or retry from an earlier point~\cite{sglang,tot,sot,selfconsistency,lmsyschat,react,autogen}. The cache must therefore support whichever boundary a future prefix match reaches, rather than only positions selected when the original prefix was processed.

For an FA-only model, this requirement follows naturally from its representation. If two requests share the first $j$ tokens, the corresponding $j$ KV entries can be reused directly at that boundary. We call this property \emph{continuous prefix caching}: reuse is available at every matched boundary supported by the cache. In a paged prefix cache, these supported positions are the page boundaries.

Hybrid models do not inherit this property directly. A token-level match identifies reusable KV entries, but the corresponding linear-attention state may be unavailable because the recurrence does not retain a separately addressable state at every earlier position. Existing systems address this mismatch by storing recurrent-state checkpoints, making a matched prefix reusable only when its boundary coincides with a stored checkpoint. Hybrid prefix caching is therefore continuous for the FA layers but discrete for the linear-attention layers.

\section{Challenges}
\label{sec:challenge}

Our goal is to provide fine-grained prefix reuse for hybrid LLMs without
sacrificing significantly the quality, storage efficiency, or serving performance of the
native prefix-caching system. Achieving this goal requires an algorithm--system
co-design. The algorithm must balance the additional state and computation
introduced by suffix replay, while the serving system must integrate these
costs with the existing KV-cache and inference pipeline.

\vspace{-0.05in}
\subsection{Algorithm level challenge: Balancing Anchor Storage, Replay Cost, and Quality}
\label{sec:challenge-algorithm}

Fine-grained prefix reuse requires reconstructing the linear-attention state at
many possible cache-hit boundaries. A naive suffix replay algorithm would retain the
hidden states needed to reconstruct the linear state at every token position. This
provides the desired reuse granularity, but its storage overhead is far beyond
the budget of our design requirement \textbf{R2}. As shown in Table~\ref{tab:naive-storage}, storing the inputs to all linear-attention layers for every token requires 20--28$\times$ more additional storage than SGLang's native hybrid cache at its default 8192-token checkpoint interval.

Reducing this storage is not an isolated compression problem. Any reduction in
the retained information changes the cost and fidelity of reconstruction.
More aggressive compression reduces storage, but may require
replaying a longer history to recover a sufficiently accurate linear-attention
state. Longer replay increases computation and can add latency on every
branching cache hit. 

The algorithmic challenge is therefore to jointly balance three coupled
quantities---the storage required by the retained replay information, the
computation required to reconstruct a state at a cache hit, and the quality of
the resulting generation. Otherwise, the system is forced to compensate for an overly expensive representation by adding more data movement or replay work to the serving critical path, violating design requirements \textbf{R1}--\textbf{R3}.

\subsection{System level challenge: Integrating Replay with the Serving Pipeline}
\label{sec:challenge-serving}

Even with an algorithmic tradeoff that meets the storage and quality targets,
suffix replay introduces state and computation that are absent from a
conventional prefix-cache hit. A hit now requires the system to obtain the
retained hidden states for replay, reconstruct the linear-attention state, and
coordinate that reconstruction with the KV cache and the forward pass that
continues the request. These operations interact with the serving system's
existing mechanisms for KV-page placement, transfer, eviction, scheduling, and
batched inference. The system challenge is therefore to incorporate replay into the native serving path without allowing its additional storage, data movement, and computation to dominate the benefits of prefix reuse. 

\begin{table}[t]
  \centering
  \small\renewcommand{\arraystretch}{1.12}%
  \caption{Additional storage per token (KiB) on top of the KV cache. Appendix~\ref{app:storage} derives both columns.}
  \label{tab:naive-storage}
  \begin{tabular}{lrrr}
    \toprule
    Model & SGlang@8192 & Naive  & Ratio \\
    \midrule
    OLMo-7B          & 6.5  & 180 & 27.6$\times$ \\
    Qwen3.5-4B       & 6.1  & 120 & 19.5$\times$ \\
    Qwen3.6-27B-FP8  & 18.4 & 480 & 26.2$\times$ \\
    \bottomrule
  \end{tabular}
\end{table}

\vspace{-0.05in}
\section{\sys{}: Algorithm Design}
\label{sec:algorithm}

\vspace{-0.02in}
The algorithmic challenge is to balance three coupled costs: the storage
required for anchors, the computation required to reconstruct a
linear-attention state, and the quality of the resulting generation. We
address this challenge by designing replay as a bounded, serving-aware
approximation rather than storing a complete state checkpoint for every
possible cache-hit position.

Our design exploits two forms of structure. First, hybrid models typically
organize linear-attention layers into groups separated by full-attention
layers. We exploit this structure to store one layer-entry anchor per group
rather than per linear-attention layer. Replay then executes only the
linear-attention computation within each group and never replays full-attention
layers, so its cost depends on the replay length rather than on the matched
context length. Finally, restarting replay from a fresh group-entry anchor
bounds approximation-error propagation to within each group, helping preserve
reconstruction quality. Second, prefix-caching systems expose cache hits at
page boundaries rather than at arbitrary token positions. Aligning anchors
with this granularity avoids storing replay inputs for unsupported hit
positions while ensuring that every supported hit boundary has a valid
reconstruction starting point.

\vspace{-0.05in}
\subsection{Layer-wise Sparse Anchors}
\label{sec:layerwise}

\begin{figure}[t]
  \centering
  \includegraphics[width=\linewidth]{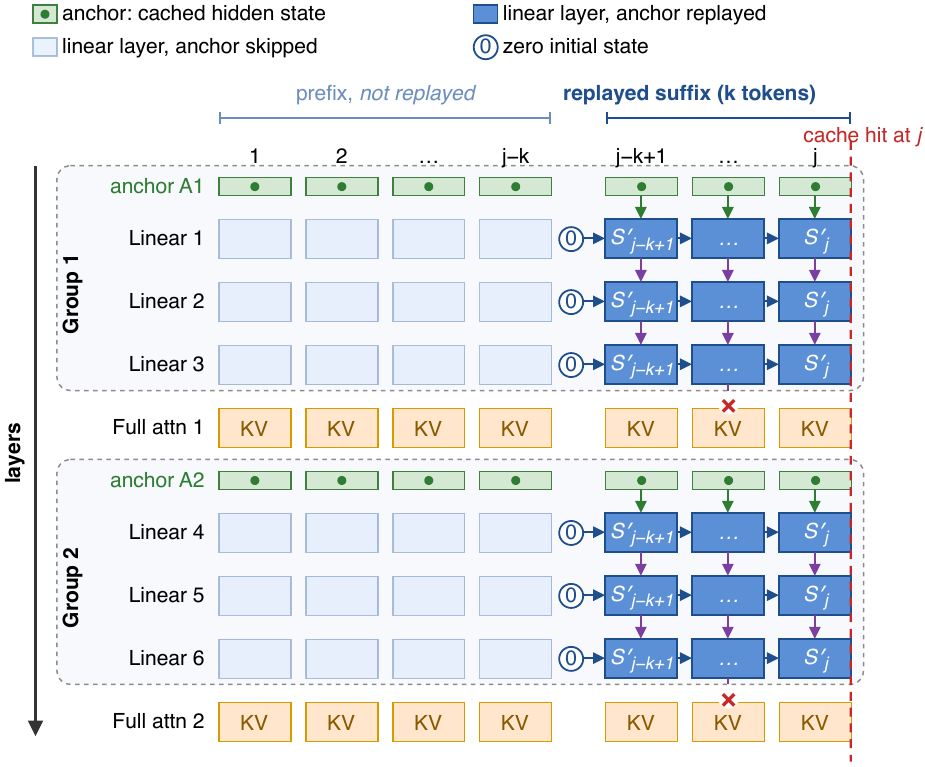}
  \vspace{-0.15in}
  \caption{Layer-wise sparse anchors. \sys{} stores the input hidden state at the entry of each group of consecutive linear-attention layers. During replay, each group reconstructs its states independently from its group-entry anchor, without crossing the intervening full-attention layers.}
  \label{fig:layerwise-sparse}
  \vspace{-0.2in}
\end{figure}

A naive representation stores the input hidden state of every linear-attention
layer at every token position. This provides accurate reconstruction at each
layer, but its storage grows with both model depth and sequence length. As
shown in Table~\ref{tab:naive-storage}, this requires 20--28$\times$ more
additional storage than SGLang's native hybrid cache across our three models.

We reduce this cost by grouping consecutive linear-attention layers into
independent replay blocks, as illustrated in Figure~\ref{fig:layerwise-sparse}.
Grouping provides three benefits. First, storing one group-entry anchor instead
of one anchor per linear-attention layer reduces the layer-wise storage
overhead from $\mathcal{O}(L)$ to $\mathcal{O}(G)$. Second, replay executes only
the linear-attention computation within each group and skips the full-attention
layers, so its computation is determined by the replay length rather than by
the length of the matched context. Third, each group restarts from its own
stored entry state. Although reconstruction errors may propagate through the
linear-attention layers within a group, they do not propagate across group
boundaries or accumulate through the full model depth.

\vspace{-0.05in}
\subsection{Token-wise Sparse Anchors}
\label{sec:tokenwise}

Storing a group-entry anchor at every token can still exceed the storage
budget. We therefore retain anchors at only a subset of token positions,
aligned with the page boundaries exposed by the prefix-cache engine. This
reduces the average storage per token while preserving the reuse granularity
supported by the serving system. With anchor density $\rho=1/16$, the
resulting storage is 0.50$\times$, 0.36$\times$, and 0.51$\times$ that of
SGLang's cache at its 8192-token checkpoint interval for OLMo-7B, Qwen3.5-4B,
and Qwen3.6-27B-FP8, respectively (Table~\ref{tab:ours-storage}).

\begin{figure}[t]
  \centering
  \includegraphics[width=\columnwidth]{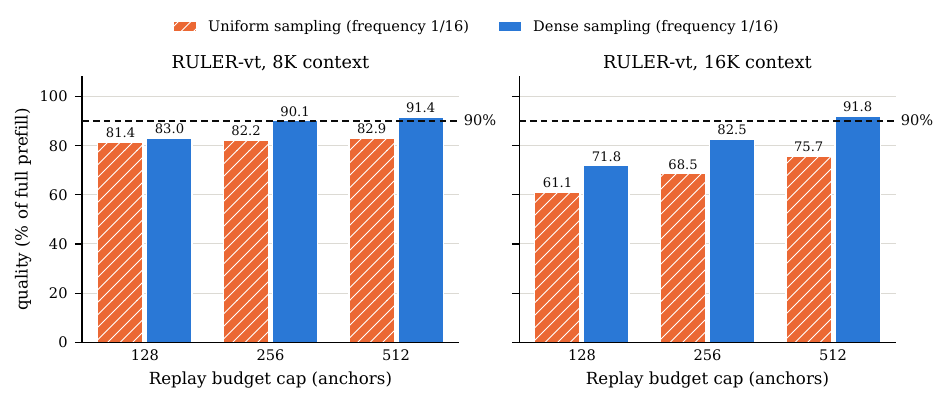}
  \vspace{-0.15in}
  \caption{Uniform (every 16th token) against dense (last four rows of every 64-token
  page) anchor sampling at the same $\rho=1/16$ frequency}
  \label{fig:dense}
  \vspace{-0.1in}
\end{figure}

The anchor layout within each page is another design choice. We compare
uniform sampling, which places anchors at regular intervals, with dense
sampling, which concentrates the retained anchors near the page boundary. Both
layouts use the same anchor density. As shown in Figure~\ref{fig:dense}, our
calibration experiments find that dense sampling provides more stable
reconstruction quality under the same replay budget and performs better on
several workloads. We therefore use the dense layout for the
experiments in the remainder of the paper.

\begin{figure}[t]
  \centering
  \includegraphics[width=\linewidth]{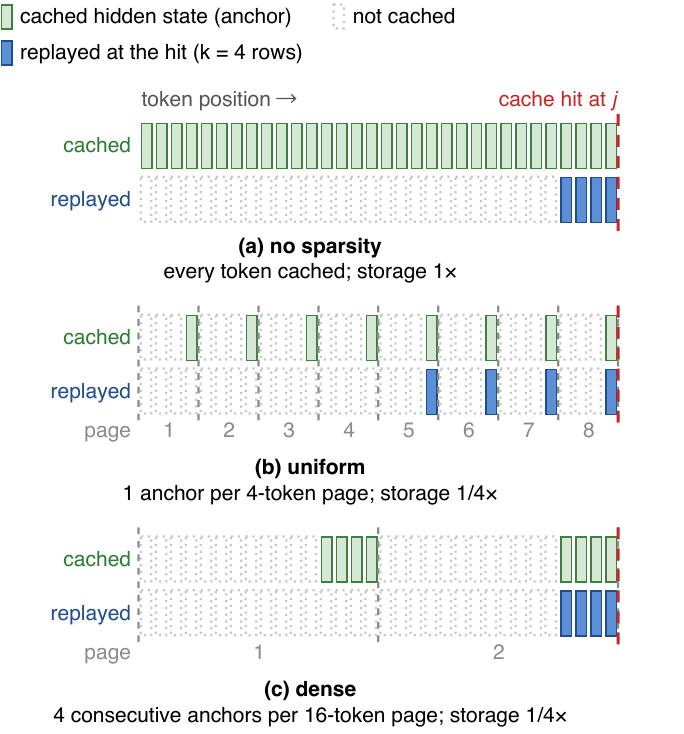}
  \vspace{-0.15in}
  \caption{Token-wise sparse anchors. No sparsity, uniform sparsity and dense sparsity}
  \label{fig:tokenwise-sparse}
  \vspace{-0.2in}
\end{figure}

Because sparse anchors cover different token spans depending on their layout,
we measure replay length in retained anchors rather than raw tokens. This
lets the algorithm bound replay computation independently of the full prefix
length. Every supported page boundary can obtain a valid starting point for
reconstruction, as illustrated in Figure~\ref{fig:tokenwise-sparse}.

\begin{table}[t]
  \centering
  \small\renewcommand{\arraystretch}{1.12}%
  \caption{Additional storage per token (KiB) with anchors at the entry of each group ($A$ layers) and anchor density $\rho=1/16$ (Ours), against SGLang's cache at its 8192-token interval and naive replay (Table~\ref{tab:naive-storage}). \sys{} stores $A \cdot d \cdot 2\rho$ bytes per token; Appendix~\ref{app:storage} gives the derivation.}
  \label{tab:ours-storage}
  \setlength{\tabcolsep}{4pt}
  \begin{tabular}{lrrrrr}
    \toprule
    Model & $A$ & SGlang & Naive & Ours & Ours/Native \\
    \midrule
    OLMo-7B          & 7  & 6.5  & 180 & 3.3 & 0.50$\times$ \\
    Qwen3.5-4B       & 7  & 6.1  & 120 & 2.2 & 0.36$\times$ \\
    Qwen3.6-27B-FP8  & 15 & 18.4 & 480 & 9.4 & 0.51$\times$ \\
    \bottomrule
  \end{tabular}
\end{table}

\vspace{-0.05in}
\subsection{Replay Length and Algorithm--System Interface}
\label{sec:replay-length}

The remaining tradeoff is controlled by the replay budget. A longer suffix generally improves the approximation to the full-prefix state but increases cache-hit computation and latency. A shorter suffix reduces replay work but may degrade generation quality. We calibrate the budget using the quality sweep( see Table~\ref{tab:quality} in section \ref{sec:eval} for detail). The table shows that both Qwen models meet the quality target at $k=128$, whereas OLMo-Hybrid-7B requires the larger budget $k=512$ on average. For a cache hit at prefix length $n$, the replay length is measured in anchors rather than tokens and follows $k=\min(0.05n,k_{\max})$. The main serving experiments use conservative caps of $k_{\max}=512$ for Qwen3.5-4B and $k_{\max}=256$ for Qwen3.6-27B-FP8 to provide additional quality headroom while bounding replay cost.

The algorithm exports a simple contract to the serving system. For each cached
page, it specifies the associated anchors; for each cache hit, it identifies
the recent anchors required for reconstruction and the corresponding
replay-length bucket. The algorithm therefore determines how much state must
be retained and how much replay is sufficient, while the system design in the
next section determines how to store, fetch, schedule, and execute this
bounded replay alongside the native KV-cache path.

\vspace{-0.1in}
\section{\sys{}: System Design}
\label{sec:system}

The algorithmic design in \S\ref{sec:algorithm} bounds the anchor storage and
replay computation required by each cache hit. It does not, however, determine
how anchors should coexist with KV-cache management or when their transfers
and reconstruction should execute relative to the serving pipeline. These are
the system-level challenges introduced by suffix replay.

Our system addresses them by decoupling logical association from physical
management. Each anchor block is associated with the KV page whose prefix it
helps reconstruct and follows the same admission and eviction decisions. Its
placement, transfer, and execution path remain independent because anchors are
read only during replay, whereas KV pages remain active throughout the
continuation. The rest of this section describes the anchor sidecar, the
pipelined replay path, and the live-state fast path for multi-turn
continuations.

\vspace{-0.05in}
\subsection{Anchor Storage and Management}
\label{sec:anchor-management}

\begin{figure}[t]
  \centering
  \includegraphics[width=\linewidth]{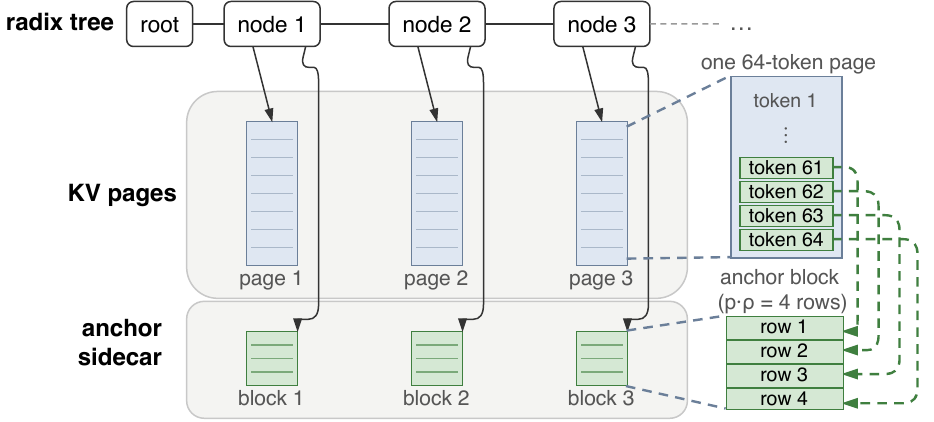}
  \vspace{-0.15in}
  \caption{The anchor sidecar: each anchor block is attached to the same radix-tree node as the KV page and shares its lifecycle, but lives in its own pool with its own placement policy.}
  \label{fig:sidecar}
  \vspace{-0.2in}
\end{figure}

A cache hit needs both the KV pages of the matched prefix and the anchors
required to reconstruct its linear-attention state, but the two objects have
different access patterns. KV pages are reused throughout the continuation,
whereas anchors are read only once during replay. A hit may also fetch a long
KV prefix but only a short suffix of anchors. Sharing a physical page would
therefore couple the KV path to data that is accessed briefly and sparsely.

We introduce an anchor sidecar to separate these concerns. As shown in
Figure~\ref{fig:sidecar}, each anchor block is attached to the same radix-tree
node as its corresponding KV page and shares its admission and eviction
decisions. The two objects, however, use independent pools and placement
policies. An anchor may reside in host DRAM while its corresponding KV page
remains in GPU HBM, without breaking prefix-cache consistency. In an HBM-only
configuration, published anchor blocks remain resident with their corresponding
KV pages. With anchor density $\rho=1/16$ and a page size of $p=64$, each page
contains $p\rho=4$ anchors, so the KV page and its anchor block are completed at
the same page boundary and require no additional alignment metadata.

Because anchors are read only during state reconstruction, a cache hit fetches
only the recent $k$ anchors specified by the algorithm, independent of the
matched prefix length. The device-side sidecar also serves as a staging area
for anchors produced by in-flight requests; sizing it proportionally to the KV
pool accounts for these unpublished blocks before they enter the shared cache
lifecycle.

\begin{figure}[t]
  \centering
  \includegraphics[width=\linewidth]{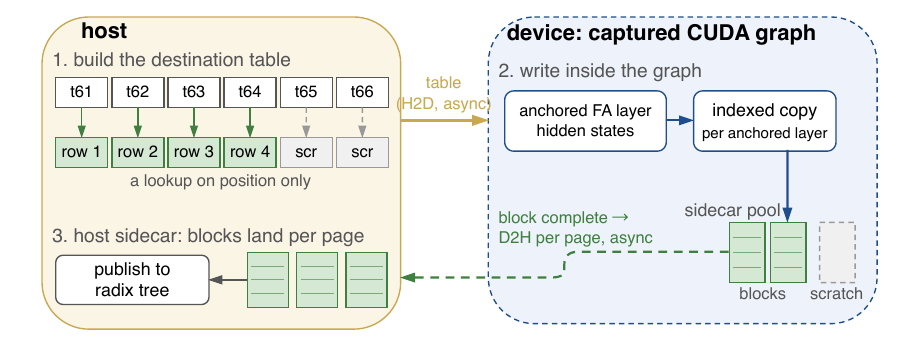}
  \vspace{-0.15in}
  \caption{Writing anchors: host builds a destination table, CUDA graph writes the hidden states.}
  \label{fig:write-path}
  \vspace{-0.2in}
\end{figure}

Anchor writes occur during every prefill and decode step, including requests
that may never become cache hits. A naive implementation copies selected
hidden states from device to host after each step and synchronizes once per
request--layer pair. With many concurrent requests, this introduces a large
number of host calls on the common serving path. In our initial implementation,
this reduced decode throughput by 3.3\% at $\rho=1/8$ and 12.1\% at
$\rho=1/2$. The bottleneck was the number of host interactions rather than the
transferred bytes.

\sys{} moves destination selection out of the forward pass and moves the
actual writes into the captured CUDA graph, as illustrated in
Figure~\ref{fig:write-path}. Because the sampling layout is static, the host
can precompute a destination table for each step. A sampled token whose page is
available receives its final sidecar address; unsampled positions receive a
scratch address and are ignored by the sidecar. The table is uploaded
asynchronously with the other inputs, and indexed copies in the CUDA graph
write the selected hidden states directly into their device-side sidecar
locations.

During prefill, \sys{} compacts the sampled tokens before the indexed copy and
flushes an anchor block only after all of its anchors have been produced.
During decode, the host updates the destination table incrementally per request
slot, while the captured graph merges writes from all anchored layers on a side
stream. Completed blocks are transferred and published asynchronously at page
granularity. On Qwen3.5-4B, a completed 64-token page transfers approximately
140\,KB. Publication is ordered after the transfer, so readers wait for a
complete block while the publisher never blocks the serving step.

\vspace{-0.05in}
\subsection{Pipelined Anchor Fetching and Replay}
\label{sec:pipelined-replay}

A cache hit introduces two data dependencies and one execution dependency
before the request can continue. The matched KV pages may need to be
transferred to the device, the anchors needed for state reconstruction must be
fetched, and replay must finish before the corresponding linear-attention group
executes. These dependencies have different readiness times, so serializing
them would unnecessarily increase hit latency.

\begin{figure}[t]
  \centering
  \includegraphics[width=\linewidth]{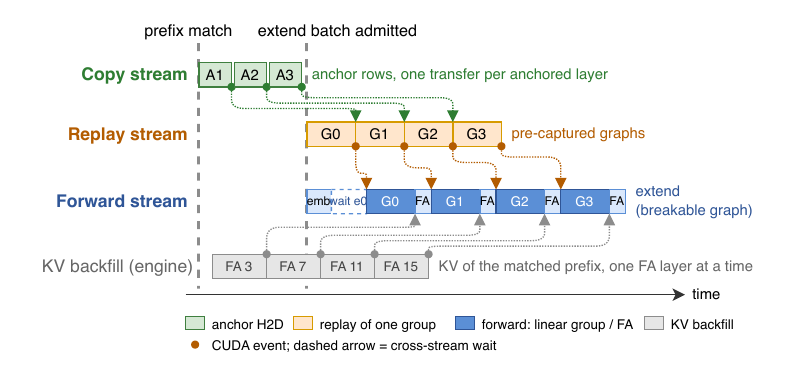}
  \vspace{-0.15in}
  \caption{Pipelined anchor fetching and replay on cache hit.}
  \label{fig:replay-pipeline}
  \vspace{-0.2in}
\end{figure}

\sys{} separates anchor fetching from replay, as shown in
Figure~\ref{fig:replay-pipeline}. As soon as the radix-tree match identifies
the cached prefix, the system walks the matched path and locates the anchor
blocks covering the replay suffix. Dense sampling places the selected anchors
contiguously near the end of each page, so the required anchors are usually
covered by a small number of ranges in the host pinned pool. \sys{} launches
their transfers asynchronously on a dedicated copy stream before KV backfill
begins. Anchor fetching therefore overlaps with request scheduling and KV
movement, and replay does not wait for the corresponding KV pages because it
depends only on the fetched anchors.

When the request enters an extend batch, \sys{} replays the linear-attention
groups from the fetched anchors. Replay executes the computation needed to
reconstruct each group and writes the resulting recurrent and convolution
states into the request's preallocated state slots. The forward stream waits
only when it reaches the first layer that consumes a group's reconstructed
state; it does not wait at request entry or between unrelated layers.

\vspace{-0.05in}
\paragraph{Layer-wise handoff.} Replay and extend execution are synchronized at
group boundaries. Groups are replayed in order on the replay stream, so replay
for later groups can remain in flight while the forward stream executes
earlier groups. Since replay length is bounded by the algorithmic policy in
\S\ref{sec:replay-length}, \sys{} pre-captures a finite set of replay graphs
indexed by replay-length buckets and groups. Runtime execution selects a graph
and fills its static input buffers, avoiding graph capture or graph eviction on
the hit path.

\vspace{-0.05in}
\paragraph{Batched replay.} Multiple hits may enter the same extend batch with
different replay lengths. Running one replay sequence per request would
serialize fixed launch and synchronization costs. \sys{} therefore packs the
replay suffixes of a batch into one variable-length execution and uses
device-side boundaries to identify each request's segment and destination state
slot. Padding segments use a sentinel slot identifier, so replay kernels
neither read nor write state for those segments. Graphs are bucketed by total
packed length, allowing one graph to serve different partitions of the same
batch. This removes repeated fixed launch costs while keeping the per-hit
replay work linear in the bounded replay length.

The pipeline is fail-closed. A non-page-aligned hit, missing or unpublished
anchors, a full pending queue, or an unavailable replay hook is treated as a
cache miss and follows the full-prefill path. A replay length outside the
captured buckets falls back to eager replay. In the steady-state hit path, host
synchronization matches the native KV-cache path, while continuation hits with
a live state bypass replay entirely.

\vspace{-0.05in}
\subsection{Multi-turn State at Request Boundaries}
\label{sec:multiturn}

A multi-turn continuation provides a special fast path. If the live
linear-attention state from the previous turn is still resident, the request
can continue without reconstructing it from anchors. This live state is only a
best-effort optimization: eviction or capacity pressure must not affect
correctness because the same state remains reconstructible from the cached KV
pages and anchors.

Requests therefore do not retain persistent checkpoint states at internal
chunk boundaries, which keeps a state pool from constraining concurrency or
cache capacity. The replay path remains the correctness path, while a live
state at a request boundary is used only when it is already available.

\section{Implementation}
\label{sec:impl}

We implement \sys{} on SGLang v0.5.18. The implementation has two parts: a small set of changes inside the engine and a separate package that provides the anchor store, replay path, and request-boundary logic. The package is attached through the engine's cache-component interface, so stock SGLang runs unchanged when the component is disabled.

\vspace{-0.05in}

\paragraph{Engine-side changes.} The engine-side implementation adds about 1.2K lines across 17 existing files and two new files. The changes cover four parts of the serving path. First, the unified radix cache gains an anchor component, allowing each anchor block to attach to the same tree node as its KV page and follow the node's admission and eviction. The same component implements the no-buffer strategy for linear states, eliminating the checkpoint pool and the extra per-request buffer slots that SGLang keeps for tracked checkpoints. Second, the pool configurator and assembler allocate the anchor sidecar, including one device block per KV page and a host-pinned pool indexed by the corresponding KV page, and account for both pools in the per-token memory budget. Third, the forward batch carries a per-token anchor-destination table, registered as a CUDA-graph input for both prefill and decode, so the in-graph writes described in \S\ref{sec:anchor-management} use the correct destinations on every graph replay. Fourth, the linear-attention operator gains a per-layer wait hook, which reduces to a single pointer comparison when replay is inactive, and the scheduler gains an asynchronous publication point for finished requests. We do not modify any attention or recurrent-state kernel.

\vspace{-0.05in}

\paragraph{Replay graphs.} \sys{} captures replay graphs once during initialization, before the engine captures its own prefill and decode graphs. For each model, the replay budget is rounded up to a power of two and capped at $K_{\mathrm{CAP}}=512$ anchors for Qwen3.5-4B and 256 anchors for Qwen3.6-27B-FP8. Production uses batched replay: graphs are bucketed by the total packed replay length, with $T_{\max}=4096$ anchors. On Qwen3.5-4B, this produces 48 graphs and approximately 1.1\,GB of device memory. All replay graphs share a memory pool because the replay stream executes them in order. Capture proceeds from smaller to larger buckets, and a memory predictor stops capture when the remaining device memory would fall below the 2.5\,GB margin; buckets that are not captured fall back to eager replay. Each graph records the base addresses of the state and anchor pools, so a pool reallocation invalidates the graph rather than allowing replay to write through stale pointers.

\section{Evaluation}
\label{sec:eval}

Our evaluation proceeds from replay quality to serving impact. First, does suffix
replay preserve end-to-end generation quality across hybrid models? Table~\ref{tab:quality}
answers this question and establishes the replay budgets used below. We then ask
whether fine-grained reuse improves serving latency under realistic branching
workloads, whether the benefit remains when the cached working set exceeds GPU
memory, and whether the high-hit continuation path avoids regressions
(Sections~\ref{sec:eval-branching}--\ref{sec:eval-capacity}). Finally, we quantify
the remaining replay and checkpoint-placement trade-offs.

\vspace{-0.05in}
\subsection{Setup}
\label{sec:setup}

\paragraph{Models and hardware.}
We evaluate replay quality on three hybrid models, OLMo-Hybrid-7B,
Qwen3.5-4B in BF16, and Qwen3.6-27B-FP8, using a PyTorch reference
implementation based on Hugging Face Transformers~\cite{transformers}, which
supports all three models. We evaluate serving performance on the two Qwen
models because the SGLang version used in our experiments does not support
OLMo-Hybrid-7B. We abbreviate Qwen3.5-4B and Qwen3.6-27B-FP8 as 4B and 27B,
respectively, when the model identity is unambiguous. All experiments use a
single NVIDIA H800 GPU with 80\,GB of
HBM. For each serving-performance cell, SGLang and \sys{} run sequentially on
the same GPU, with the execution order alternated across repeated runs.

\vspace{-0.05in}

\paragraph{Engine and baseline.}
Both systems use the same SGLang checkout and the same scheduler, KV-cache
implementation, 8192-token chunked prefill, and CUDA-graph configuration. The
baseline is SGLang's native hybrid prefix cache without modification. Under its
default configuration, SGLang stores linear-attention state checkpoints at
prefill-chunk boundaries, at the end of a request (aligned down to 256 tokens),
and when a cached prefix first serves a branching continuation. These checkpoints are kept in a fixed
HBM slot pool and are required to resume a hybrid request at the corresponding
boundary. \sys{} replaces this checkpoint store with the sparse-anchor sidecar
and suffix-replay path described in Sections~\ref{sec:algorithm} and
\ref{sec:system}; all other serving components remain unchanged.

\vspace{-0.05in}

\paragraph{Configuration.}
Unless a caption says otherwise, \sys{} uses dense anchors with $\rho=1/16$ on
64-token pages, retaining the last four anchor rows of each page. For a cache hit
at prefix length $n$, the replay budget is $k=\min(0.05n,k_{\max})$ anchor rows.
The quality experiments sweep $k$ over multiple budgets; Section~\ref{sec:eval-quality}
reports the resulting quality curves. The main serving experiments use the
conservative caps $k_{\max}=512$ for Qwen3.5-4B and $k_{\max}=256$ for
Qwen3.6-27B-FP8, providing quality headroom while bounding replay cost on the
larger model. These caps were fixed before the quality sweep completed;
Section~\ref{sec:eval-sensitivity} evaluates the lower-cost alternatives.

\vspace{-0.05in}

\paragraph{Memory configuration.}
Unless stated otherwise, experiments use an HBM-only configuration: both systems
keep their KV cache and linear states on the GPU, and \sys{} retains anchors only
for KV pages that remain resident. The capacity experiments enable SGLang's host
KV cache for both systems under a fixed 128\,GB host-memory budget. SGLang
uses approximately 128\,GB for host KV pages, whereas \sys{} reserves up to
8\,GB of pinned host memory for the anchor sidecar: 4.7\,GB in the capacity
experiment and 8\,GB when the anchor pool is sized to the host KV cache in
LooGLE. The remaining host memory is used for KV pages. Thus, the two systems have comparable
total host-memory budgets, but only \sys{} can move replay metadata together
with a KV page; SGLang's linear-state checkpoints remain in its fixed HBM slot
pool.

\vspace{-0.05in}

\paragraph{Workloads.}
We use \emph{branch point} to denote the first position at which a new request
diverges from a previously cached prefix; the tokens before this position form
the matched prefix. We evaluate four workload groups. For branching reuse, we use \emph{ShareGPT}
multi-turn chat sessions~\cite{lmsyschat}, whose turns extend the same context;
\emph{SWE-rebench}, OpenHands agent trajectories with long tool-call sessions
and short model outputs; and \emph{ToolMind}, tool-use trajectories that branch
from a point inside a previous response. The median ToolMind continuation point
is approximately 5K tokens; we also report a long-context variant restricted to
continuations beyond 16K tokens. For document question answering, we use
\emph{LooGLE}~\cite{loogle}, whose documents contain approximately 40K tokens,
and \emph{NarrativeQA}~\cite{narrativeqa}, whose documents are shorter and whose
questions in our trace branch at the document boundary. We use \emph{Agent
continuation}, a set of 240 multi-turn agent traces with short outputs and a
98\% prefix-hit rate, to evaluate the high-hit regime. Finally, two synthetic
workloads stress capacity and checkpoint placement: $N$ sessions repeatedly
query independent 64K-token documents, and a shared document prefix is
re-queried at shrinking and growing prefix lengths to probe both sides of the
native checkpoint grid.

\vspace{-0.05in}

\paragraph{Protocol and metrics.}
Serving workloads use either open-loop or closed-loop evaluation. In open-loop
runs, requests arrive according to a Poisson process at rate $\lambda$, measured
per session for session traces and per request otherwise. In closed-loop runs,
we keep a fixed number $N$ of sessions in flight. Unless a caption states
otherwise, output lengths are fixed by the workload and range from 64 to 512
tokens. We report client-observed time to first token (TTFT), time per output token (TPOT),
request throughput, and prefix-cache hit rate. We count a request as a hit when
the cache returns the prefix through its intended continuation point.

\vspace{-0.05in}
\subsection{Quality}
\label{sec:eval-quality}

We first verify that suffix replay is a valid proxy for full prefill. For each
sample, we split the input into a context prompt and a query. We prefill the
context once, retain its FA KV cache and sparse anchor inputs, and then emulate a
cache hit at the query boundary: suffix replay reconstructs the linear-attention
states, the retained FA KV cache is reused, and the query is run through the
remaining forward pass. We compare this output with a paired reference that
full-prefills the context and query together. Table~\ref{tab:quality} reports
the resulting replay quality for all three hybrid models across LongBench and
RULER as the replay budget varies. The models exhibit markedly different
sensitivity to replay truncation. OLMo-Hybrid-7B requires a larger budget: at
$k=512$, its family averages reach 95.9\% of full-prefill quality on LongBench
QA and 91.4\% on RULER. In contrast, the two Qwen models are nearly insensitive
to replay truncation. At $k=128$, Qwen3.5-4B reaches 100\% and 99.7\% on QA
and RULER, respectively, while Qwen3.6-27B-FP8 reaches 99.1\% and 100\%.
Even at $k=16$, the two models remain at 95.6/94.2\% and 99.0/99.4\%,
respectively. Thus, $k=128$ provides substantial quality headroom for both
Qwen models, whereas OLMo requires the larger replay budget.

We use each benchmark's standard scorer: token-level F1 for LongBench QA,
ROUGE-L for LongBench summarization, and RULER's exact string-match metrics
(all-reference or any-reference matching according to the task). We report
$\%\mathrm{full}=100\times s_{\mathrm{replay}}/s_{\mathrm{full}}$, where both
scores are computed on the same samples and $s_{\mathrm{full}}$ is obtained by
the paired full-prefill reference. Thus, the reported ratio isolates the quality
effect of reconstructing the linear-attention states.

\begin{table*}[t]
  \centering
  \normalsize
  \renewcommand{\arraystretch}{1.0}%
  \caption{Anchor replay quality relative to full prefill (\%full) for different
  replay budgets $k$. Each entry is a paired sample average; entries below 90\%
  are bold. @ denotes context length.}
  \label{tab:quality}
  \setlength{\tabcolsep}{2.4pt}
\begin{tabular}{lrrrrrr@{\hspace{9pt}}rrrr@{\hspace{9pt}}rrrr}
  \toprule
  & \multicolumn{6}{c}{OLMo-Hybrid-7B} & \multicolumn{4}{c}{Qwen3.5-4B} & \multicolumn{4}{c}{Qwen3.6-27B-FP8} \\
  \cmidrule(lr){2-7}\cmidrule(lr){8-11}\cmidrule(lr){12-15}
  Cell & $k{=}16$ & 32 & 64 & 128 & 256 & 512 & $k{=}16$ & 32 & 64 & 128 & $k{=}16$ & 32 & 64 & 128 \\
  \midrule
\multicolumn{15}{l}{\textit{LongBench QA}}\\
\quad NarrativeQA@16K & \textbf{85.8} & \textbf{88.1} & 91.1 & 92.5 & 93.6 & 97.2 & 97.5 & 98.5 & 100 & 100 & 98.3 & 97.2 & 100 & 99.3 \\
\quad NarrativeQA@32K & \textbf{84.2} & 91.0 & 93.3 & 94.8 & 95.3 & 95.4 & 91.6 & 91.8 & 97.4 & 97.9 & 100 & 95.8 & 99.2 & 100 \\
\quad NarrativeQA@64K & \textbf{85.6} & \textbf{82.9} & \textbf{86.8} & 92.8 & 93.3 & 96.8 & 94.1 & 96.5 & 100 & 99.2 & 95.7 & 100 & 97.2 & 99.7 \\
\quad HotpotQA@16K & \textbf{86.8} & \textbf{89.9} & 90.9 & 95.5 & 92.5 & 94.1 & 99.2 & 99.5 & 99.6 & 100 & 99.2 & 98.1 & 97.5 & 97.4 \\
\addlinespace[2pt]
\quad GovReport@32K & 96.7 & 97.0 & 97.1 & 97.0 & 96.8 & 97.6 & 95.0 & 95.3 & 96.0 & 94.8 & 97.3 & 96.5 & 97.4 & 96.7 \\
\addlinespace[2pt]
\multicolumn{15}{l}{\textit{RULER}}\\
\quad NIAH@8K & \textbf{88.7} & 90.1 & 92.3 & 92.6 & 91.8 & 90.9 & 99.1 & 99.1 & 99.4 & 100 & 100 & 100 & 100 & 100 \\
\quad NIAH@16K & 92.0 & 92.1 & \textbf{89.9} & 90.0 & \textbf{89.1} & 91.5 & 97.6 & 97.6 & 97.9 & 100 & 100 & 100 & 100 & 100 \\
\quad CWE@8K & \textbf{85.0} & \textbf{85.9} & \textbf{87.0} & \textbf{88.2} & 91.1 & 90.1 & 99.5 & 99.6 & 99.7 & 100 & 100 & 100 & 100 & 100 \\
\quad CWE@16K & \textbf{88.1} & \textbf{87.5} & \textbf{89.4} & \textbf{88.8} & 90.0 & 95.2 & 98.5 & 100 & 100 & 100 & 100 & 100 & 100 & 100 \\
\quad QA-1@8K & \textbf{80.0} & \textbf{85.4} & \textbf{87.5} & 93.8 & 93.1 & 91.7 & 95.3 & 94.2 & 96.5 & 99.4 & 98.3 & 98.9 & 99.4 & 100 \\
\quad QA-1@16K & \textbf{68.6} & \textbf{68.6} & \textbf{74.3} & \textbf{81.4} & \textbf{84.9} & \textbf{89.3} & \textbf{75.5} & \textbf{78.4} & \textbf{88.0} & 98.2 & 98.3 & 99.4 & 100 & 100 \\
\quad VT@8K & \textbf{68.9} & \textbf{74.4} & \textbf{80.0} & \textbf{84.2} & 90.1 & 92.2 & 100 & 100 & 100 & 100 & 99.9 & 99.9 & 99.9 & 99.9 \\
\quad VT@16K & \textbf{38.5} & \textbf{48.7} & \textbf{60.6} & \textbf{71.9} & \textbf{82.5} & 91.8 & 100 & 100 & 100 & 100 & 100 & 100 & 100 & 100 \\
\quad VT@32K & \textbf{75.8} & \textbf{80.5} & \textbf{80.9} & \textbf{86.4} & \textbf{89.0} & 91.5 & 98.3 & 98.7 & 99.0 & 98.7 & 100 & 100 & 100 & 100 \\
\quad VT@64K & \textbf{67.2} & \textbf{66.8} & \textbf{76.3} & \textbf{87.5} & \textbf{89.9} & 96.6 & 99.0 & 99.0 & 99.3 & 99.3 & 100 & 100 & 100 & 100 \\
\quad VT@128K & \textbf{41.1} & \textbf{49.8} & \textbf{58.9} & \textbf{71.1} & \textbf{78.2} & \textbf{81.7} & 90.3 & \textbf{88.3} & \textbf{89.9} & 93.3 & 100 & 100 & 100 & 100 \\
\addlinespace[2pt]
\midrule
\textit{Avg LongBench QA} & \textbf{85.6} & \textbf{88.0} & 90.5 & 93.9 & 93.7 & 95.9 & 95.6 & 96.6 & 100 & 100 & 99.0 & 98.1 & 99.1 & 99.1 \\
\textit{Avg RULER} & \textbf{83.7} & \textbf{84.9} & \textbf{86.7} & \textbf{89.2} & \textbf{90.0} & 91.4 & 94.2 & 94.8 & 96.9 & 99.7 & 99.4 & 99.7 & 99.9 & 100 \\
  \bottomrule
\end{tabular}

  \renewcommand{\arraystretch}{1.0}
\end{table*}

\subsection{Branching prefix reuse}
\label{sec:eval-branching}

\begin{figure*}[t]
  \centering
  \includegraphics[width=0.95\textwidth]{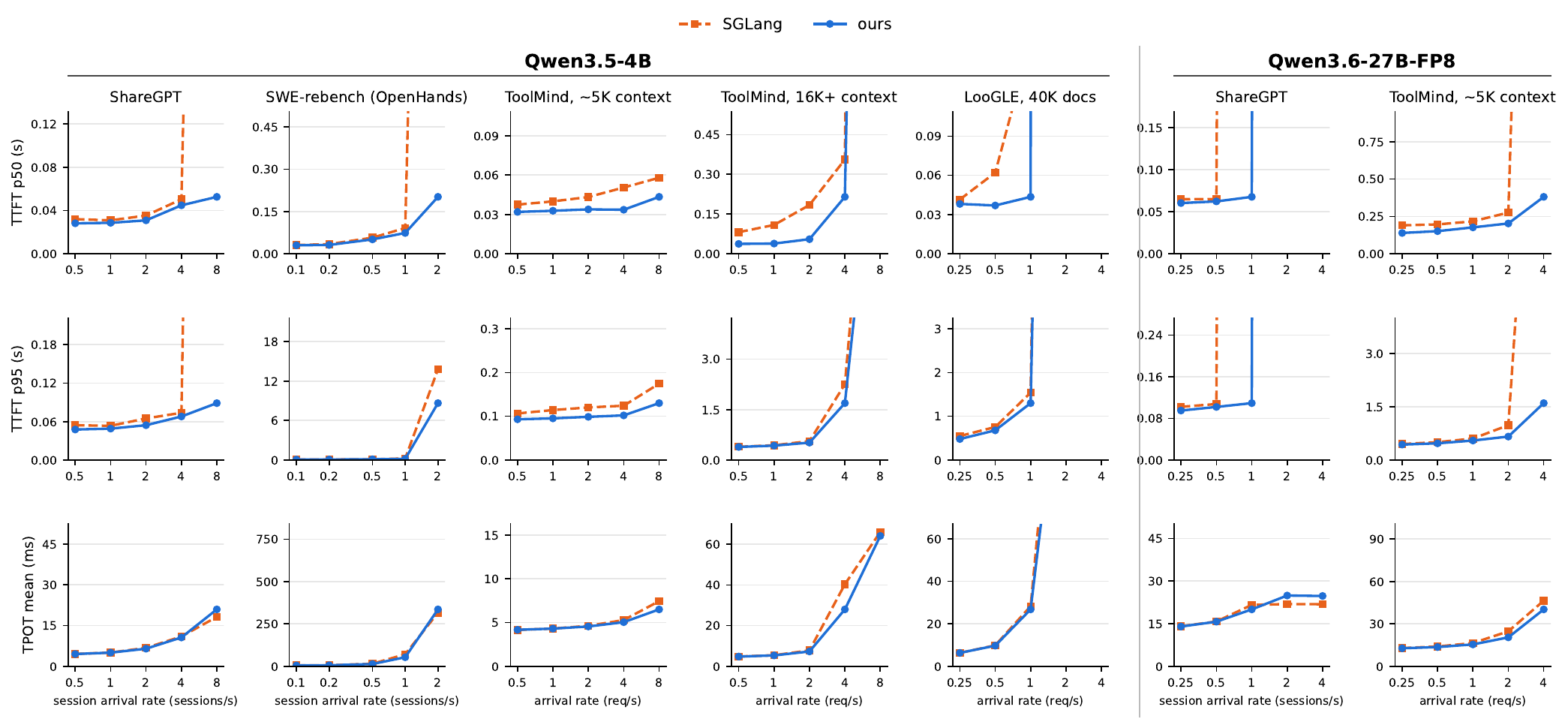}
  \vspace{-0.25in}
  \caption{TTFT p50/p95 and mean TPOT versus offered load on the evaluated
  workloads. All runs use HBM-only serving; session workloads use session arrival
  rate and other workloads use request arrival rate.}
  \label{fig:lam}

\end{figure*}

\paragraph{Branching workloads across arrival rates.}
Figure~\ref{fig:lam} shows TTFT and TPOT as the offered load increases. On every
branching workload, \sys{} lowers median TTFT: by 15--20\,\% on ShareGPT,
20--40\,\% on ToolMind at approximately 5K context, and 55--70\,\% on ToolMind
beyond 16K. The gain is largest when the branch point falls far beyond SGLang's
latest checkpoint. At high load, SGLang saturates earlier because its tracked
linear-state slot pool limits admission: ShareGPT saturates at 4 sessions/s on
4B and 0.5 sessions/s on 27B, while \sys{} remains stable at 8 and 1
sessions/s, respectively. TPOT changes little: although every decode step
produces anchor states, the CUDA-graph write path places them into the sidecar
without host synchronization (\S\ref{sec:anchor-management}), so anchor writes
add negligible decode overhead.

\vspace{-0.05in}

\begin{figure}[t]
  \centering
  \includegraphics[width=\columnwidth]{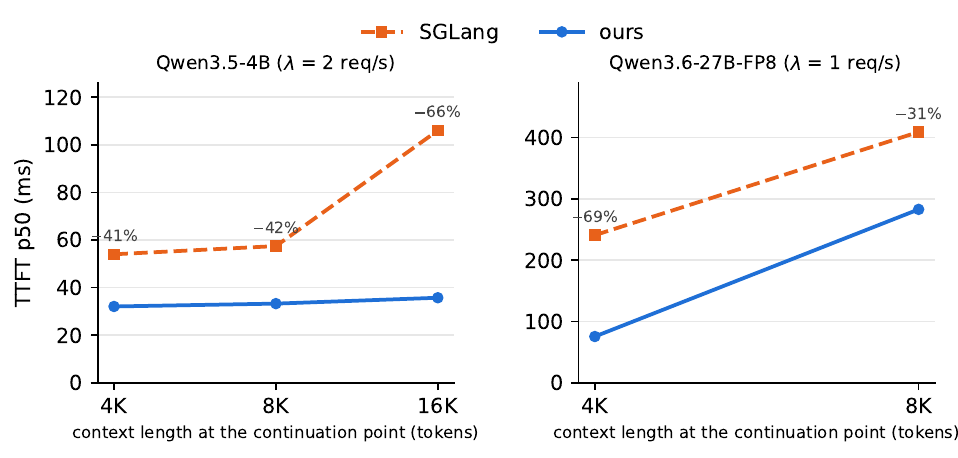}
  \vspace{-0.15in}
  \caption{Median TTFT of ToolMind continuations versus branch context length.
  Results use open-loop, HBM-only serving.}
  \label{fig:ctx}
 \vspace{-0.15in}
\end{figure}

\paragraph{Distance from the checkpoint grid.}
The benefit grows with the distance between the branch point and SGLang's latest
checkpoint. Figure~\ref{fig:ctx} varies the continuation context on ToolMind.
On 4B, \sys{} reduces median TTFT by 41--42\,\% at 4K--8K context and by
66\,\% at 16K, where the continuation falls 3--4K tokens beyond a checkpoint.
On 27B, the reductions are 69\,\% at 4K and 31\,\% at 8K. Shrinking SGLang's
prefill chunk size directly shortens its checkpoint interval and improves reuse
granularity. However, more frequent checkpoints increase
cold-prefill latency and amortized checkpoint storage. Table~\ref{tab:chunk}
quantifies this tradeoff.

\vspace{-0.05in}

\begin{figure}[t]
  \centering
  \includegraphics[width=\columnwidth]{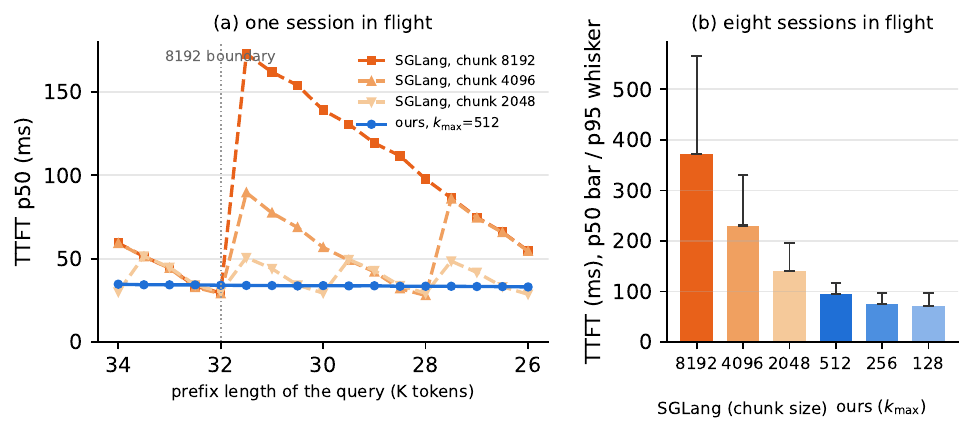}
  \vspace{-0.15in}
  \caption{TTFT for branches from a shared 35K-token prefix at different cut
  positions. Panel (a) shows p50 TTFT with one session in flight; panel (b)
  shows p50 with p95 whiskers under eight concurrent sessions. SGLang uses
  8192-, 4096-, or 2048-token chunks, while \sys{} uses the corresponding
  replay caps.}
  \label{fig:grid}

\end{figure}

\paragraph{Branches between checkpoints.}
Figure~\ref{fig:grid} constructs the case that is unfavorable to SGLang:
requests branch from the same long prefix at cut positions between its
checkpoints. SGLang then recomputes 0.6--7.7K tokens, producing a saw-tooth TTFT
from 30 to 175\,ms, whereas \sys{} serves every cut in about 33\,ms from the
corresponding page anchors. With eight sessions in flight, \sys{} reduces median
TTFT from 369\,ms to 72--94\,ms ($3.9$--$5.1\times$); even after SGLang's
prefill chunk is reduced to 2048 tokens, thereby shortening its checkpoint
interval, SGLang remains $1.5$--$2\times$ slower.

\begin{figure}[t]
  \centering
  \includegraphics[width=\columnwidth]{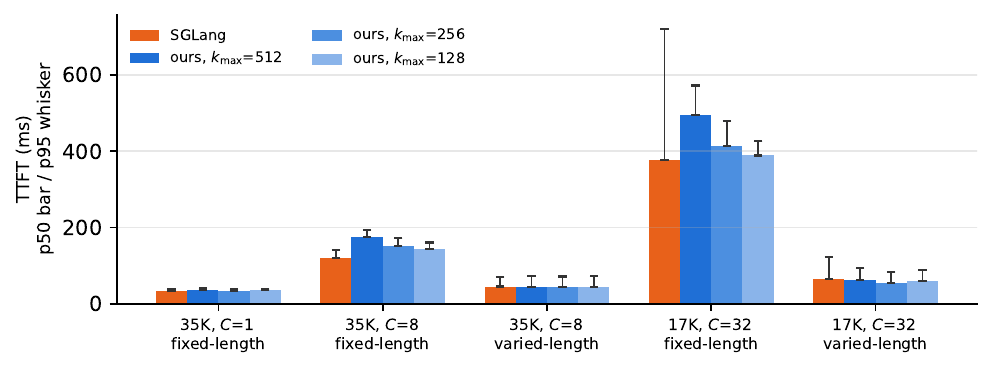}
  \vspace{-0.15in}
  \caption{TTFT when querying the same prefix at growing lengths. Bars show p50
  and whiskers show p95 for Qwen3.5-4B in HBM-only serving; results compare
  fixed-length and varied-length outputs.}
 \vspace{-0.15in}
  \label{fig:grow}

\end{figure}

Figure~\ref{fig:grow} provides the opposite, SGLang-favorable case: every branch
point coincides with the previous request's end checkpoint, so SGLang resumes
without recomputation while \sys{} still replays its suffix. With one session in
flight, the 35K-token runs differ by at most 2\,ms. Fixed-length outputs make
multiple sessions finish and re-branch in lock-step, exposing the synchronized
cache-hit and request-boundary work: at 35K tokens and eight sessions, the
medians are 121\,ms for SGLang and 174/152/143\,ms for \sys{} with
$k_{\max}=512/256/128$; at 17K tokens and 32 sessions, they are 377\,ms and
494/413/389\,ms, respectively. Assigning varied output lengths from 16--112
tokens breaks this synchrony and removes the gap: the corresponding medians are
45\,ms versus 45/43/43\,ms at 35K and 65\,ms versus 62/54/59\,ms at 17K, with
a lower p95 for \sys{}. We analyze this synchronized case and its overhead in
Section~\ref{sec:eval-overhead}; the paired workloads here show that \sys{}
removes SGLang's checkpoint-alignment penalty while remaining close to SGLang's
best case.

\vspace{-0.08in}
\subsection{High-hit continuation}
\label{sec:eval-cont}

Does the anchor path cost anything when SGLang already hits? Agent traces
are the adversarial case: 98\,\% of requests continue the previous turn of their
session, exactly where SGLang keeps an end-of-request checkpoint, so the
SGLang cache copies a slot and \sys{} must not be slower. Table~\ref{tab:continuation}
shows it is not: a continuation hit hands the previous turn's live state over
without replay (Section~\ref{sec:system}), and \sys{} matches or slightly exceeds the
SGLang throughput at every concurrency, $+2.7$ to $+6.5$\,\% at 8 and 16 sessions
and a tie ($-1.4$\,\%, within round-to-round noise) at 32 on 4B, with TTFT
equal or lower. The small gains come from the pages SGLang's grid misses: an
anchor hit also covers turns that resume off the 256-token alignment of the SGLang
end checkpoint.

\begin{table}[t]
  \centering
  \small\renewcommand{\arraystretch}{1.12}%
  \caption{High-hit agent continuation with 240 multi-turn traces and a 98\% hit
  rate. Results use closed-loop HBM-only serving; $N$ is the number of concurrent
  sessions.}
  \label{tab:continuation}
  \begin{tabular}{@{}llrrr@{}}
    \toprule
    & & \multicolumn{2}{c}{throughput (req/s)} & TTFT p50 (ms) \\
    \cmidrule(lr){3-4}
    model & $N$ & SGLang & ours & SGLang $\to$ ours \\
    \midrule
    Qwen3.5-4B & 8  & 21.9 & 23.1 (+5.4\,\%) & 54 $\to$ 47 \\
               & 16 & 26.9 & 27.6 (+2.7\,\%) & 62 $\to$ 57 \\
               & 32 & 32.7 & 32.2 ($-1.4$\,\%) & 68 $\to$ 65 \\
    Qwen3.6-27B-FP8 & 8  & 8.8 & 9.3 (+5.6\,\%) & 97 $\to$ 86 \\
               & 16 & 11.4 & 12.1 (+6.5\,\%) & 119 $\to$ 107 \\
    \bottomrule
  \end{tabular}
\end{table}

\vspace{-0.05in}
\subsection{Capacity}
\label{sec:eval-capacity}

\begin{figure}[t]
  \centering
  \includegraphics[width=\columnwidth]{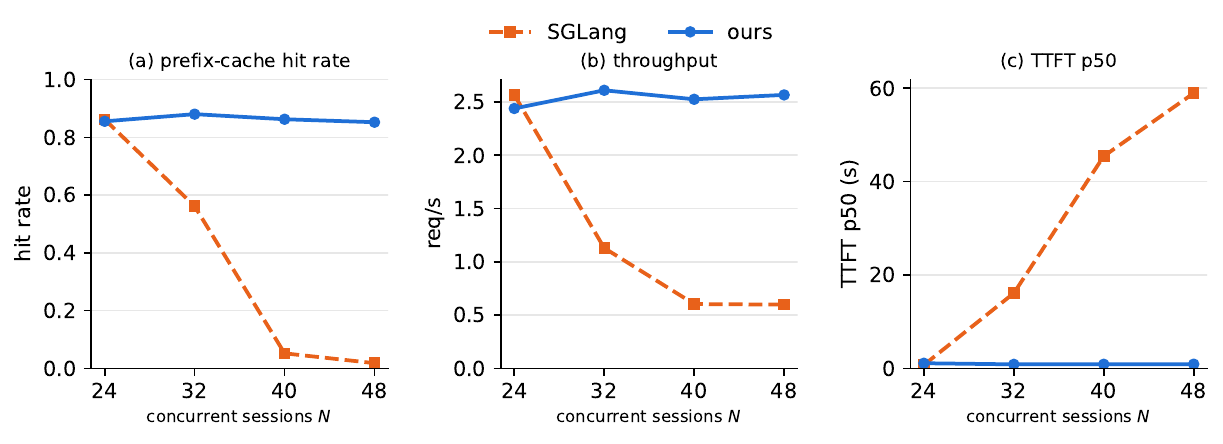}
  \vspace{-0.15in}
  \caption{Capacity under $N$ sessions querying independent 64K-token documents
  with host KV caching. Panels show prefix-cache hit rate, throughput, and median
  TTFT.}
  \label{fig:capacity}
  
\end{figure}

What happens when the cached working set outgrows HBM? SGLang keeps its
linear states in a fixed slot pool on the GPU; \sys{}'s anchors are ordinary pages
that move with the KV cache. Figure~\ref{fig:capacity} drives $N$ sessions that
each keep querying their own 64K-token document with the host KV cache enabled.
Below the cliff ($N=24$, 1.5M tokens) the two are on par, \sys{} 5\,\% lower in
throughput. From $N=32$ the documents no longer fit the 983K-token device pool.
SGLang still holds every document in its radix tree -- on every miss the
tree matches the full 64K prefix -- but under this pressure it brings a document
back from the host for only a small fraction of the requests (87 load-backs
against 668 cold prefills in an instrumented run) and recomputes the rest, so its
hit rate falls from 0.86 to 0.05 and throughput from 2.6 to 0.6\,req/s; \sys{}
stays at 0.85 and 2.5--2.6\,req/s, $2.3$--$4.3\times$ SGLang's throughput, with a
median TTFT under one second against 16--59\,s. SGLang's slot pool is not the
limit (peak occupancy 11\,\%); the limit is that its states do not page.

A natural document-QA load, LooGLE, has 4.6M tokens of context and exceeds the
device pool several times over. At
0.5--2\,req/s \sys{} answers a follow-up question in 53--159\,ms at the median
against 426--1,307\,ms, with a hit rate of 0.62 against 0.35; at 4\,req/s SGLang saturates and its queue grows for the whole run, while \sys{} still serves
the offered load. The anchor host pool that makes this possible is one block per
host KV page, 8\,GB of pinned memory or 7\,\% of the host KV cache.

\subsection{Replay-length sensitivity}
\label{sec:eval-sensitivity}

\paragraph{How much to replay.}
The replay budget controls the cost--quality trade-off. Table~\ref{tab:quality}
shows that both Qwen models reach the quality target at $k=128$, while OLMo
requires a larger budget on average. The main serving experiments nevertheless
use the conservative $k_{\max}=512$ (4B) and $256$ (27B) caps, which were fixed
before the quality sweep completed; $k_{\max}=128$ is the recommended lower-cost
setting when additional quality headroom is not required.

\subsection{Replay overhead and checkpoint trade-offs}
\label{sec:eval-overhead}

\paragraph{What SGLang's grid would cost.}
SGLang can only place checkpoints more often by shrinking its prefill
chunk. Table~\ref{tab:chunk} prices that knob: chunks of 2048 tokens raise the cold
prefill TTFT by 22\,\% and quadruple the state bytes per cached token, chunks of
1024 raise it by 87\,\% and cost 16\,\% of closed-loop throughput, and even the
2048-token grid is still $1.5$--$2\times$ slower than \sys{} on the shrinking-prefix
workload (Figure~\ref{fig:grid}).

\begin{table}[t]
  \centering
  \footnotesize
  \caption{Effect of smaller SGLang prefill chunks on cold-prefill TTFT, throughput,
  and checkpoint storage. Results use Qwen3.5-4B in HBM-only serving.}
  \label{tab:chunk}
  \begin{tabular}{@{}lrrr@{}}
    \toprule
    SGLang prefill chunk & 8192 (stock) & 2048 & 1024 \\
    \midrule
    cold prefill TTFT p50 (ms) & 677 & 823 (+22\,\%) & 1,268 (+87\,\%) \\
    throughput (req/s) & 13.8 & 14.1 & 11.5 ($-16$\,\%) \\
    state bytes per cached token & 6,286 & 25,144 & 50,288 \\
    \bottomrule
  \end{tabular}
\end{table}

\FloatBarrier  %

\section{Related Work}
\label{sec:related}

\paragraph{Prefix caching for Transformers.}
Because a Transformer's KV cache is indexed by token, any prefix of a cached
sequence is reusable, and prefix-caching systems concentrate on where to keep the
pages and how to schedule around them: paged allocation and radix-tree
matching~\cite{pagedattention,sglang}, session-level reuse across
turns~\cite{cachedattention}, disaggregated and tiered KV
stores~\cite{mooncake}, and cache-aware request routing~\cite{preble}. All of
them assume that a cached prefix can be cut at any position. Hybrid models
break that assumption for their linear layers, which is the gap this paper fills.

\paragraph{Position-independent KV reuse.}
To reuse cached KV beyond exact prefixes, PromptCache, CacheBlend, EPIC and
ProphetKV~\cite{promptcache,cacheblend,epic,prophetkv} accept KV computed in a
different context and recompute a small, selected set of tokens to repair the
rest; they target Transformers and say so. HyPIC and
LinearKV~\cite{hypic,liu2026linearkvcachedstatesuffices} extend
position-independent caching to hybrid models.

\paragraph{Prefix caching for hybrid and recurrent models.}
Serving engines cache linear-layer states at checkpoints: SGLang keeps one per
prefill chunk and per request end~\cite{sglang}, and vLLM's automatic prefix
caching for hybrid models is under active
discussion~\cite{vllmapc,vllm40696,vllm45238,vllm46384}. Marconi~\cite{marconi}
decides which checkpoints to admit and evict, and SparsePrefix~\cite{sparseprefix}
thins the set of states kept. All of these store states and therefore inherit
the question of where to place them; \sys{} stores inputs and reconstructs the
state where the hit happens, so the placement question does not arise, and
admission policies such as Marconi's remain useful for deciding which anchor
pages to retain.

\paragraph{Linear attention and forgetting.}
Mamba-2, Gated DeltaNet and the hybrid models built on them~\cite{mamba2,
gateddeltanet,jamba,samba,nemotronh,minimax01,qwen35,qwen36,kimilinear,olmohybrid}
share the decay and erase gates that make a state a contracting function of its
distant inputs. That property is what suffix replay relies on
(Section~\ref{sec:algorithm}); recurrences without decay are outside its scope.

\paragraph{AI serving across heterogeneous resources.}
Recent work on AI Flow studies how to coordinate heterogeneous computing and
communication resources across devices, edge servers, and cloud infrastructure
to reduce inference latency and communication overhead~\cite{aiflowedge,aiflowperspectives}.
Our work focuses on a complementary intra-engine problem: enabling fine-grained
prefix reuse for hybrid LLMs by reconstructing linear-attention states while
coordinating anchors, KV caches, and replay within a serving system.

\section{Limitations}
\label{sec:discussion}

\sys{} relies on linear-attention states progressively forgetting earlier inputs and therefore does not directly apply to recurrences that preserve all past contributions. The required replay budget is also model dependent: the Qwen models meet the quality target with $k=128$, whereas OLMo requires a larger budget. Our current implementation calibrates this budget offline for each model.

Replay can increase TTFT when many cache hits arrive in the same scheduler step, particularly when SGLang already has checkpoints at the matched boundaries. Figure~\ref{fig:grow} shows this overhead under synchronized, fixed-length requests; varying the output lengths removes the gap.
\section{Conclusion}
\label{sec:conclusion}

Prefix caching for hybrid LLMs has been discrete because the linear layers'
state is stored at checkpoints and a cached prefix can only be resumed where a
checkpoint exists. \sys{} removes the checkpoints: it stores a sparse set of the
linear layers' inputs as anchors and rebuilds the state at any 64-token page
boundary by replaying a short suffix, relying on the recurrence's own forgetting
to make the distant past irrelevant. Integrated into SGLang, it cuts the median
TTFT of continuations branching from shared prefixes by 41--69\,\% and sustains twice the arrival rate of
SGLang's cache, keeps its hit rate and $2.3$--$4.3\times$ the throughput when
the working set outgrows HBM, ties SGLang's cache on high-hit continuation
traffic, retains at least 94\,\% of full-recompute quality, and stores
$0.36$--$0.51\times$ the state bytes -- all while making every page boundary
reusable.

\bibliographystyle{ACM-Reference-Format}
\bibliography{references}

\appendix

\section{Per-token storage accounting}
\label{app:storage}

This appendix derives the storage numbers reported in Tables~\ref{tab:naive-storage} and~\ref{tab:ours-storage}. Both columns
count storage \emph{in addition to} the KV cache of the full-attention layers,
which the two designs store identically. All sizes are in bytes; 1\,KiB = 1024
bytes.

\vspace{-0.05in}

\paragraph{Native cache.}
A checkpoint of SGLang's cache holds, for each of the $L_{\text{lin}}$
linear-attention layers, the two pieces of state that the layer carries across
tokens. The first is the recurrent state, one $d_k \times d_v$ matrix per value
head, kept in FP32:
\begin{equation}
  B_{\text{rec}} = H_v \cdot d_k \cdot d_v \cdot 4 .
\end{equation}
The second is the state of the short causal convolution applied to the query,
key and value projections. With kernel width $w$, the layer keeps the last
$w-1$ inputs of each of its $C = 2H_k d_k + H_v d_v$ channels, in BF16:
\begin{equation}
  B_{\text{conv}} = C \cdot (w-1) \cdot 2 .
\end{equation}
One checkpoint therefore occupies
$B_{\text{ckpt}} = L_{\text{lin}}\,(B_{\text{rec}} + B_{\text{conv}})$ bytes.
The native cache takes a checkpoint every $T$ tokens, so its amortized cost is
\begin{equation}
  B_{\text{native}}(T) = B_{\text{ckpt}} / T
  \label{eq:native-bpt}
\end{equation}
bytes per token, with $T = 8192$ by default. The cost is inversely
proportional to $T$: halving the checkpoint interval doubles it.

\vspace{-0.05in}

\paragraph{Naive replay.}
Storing the input of every linear-attention layer at every token keeps one
hidden-state vector of width $d$ per layer per token, in BF16:
\begin{equation}
  B_{\text{naive}} = L_{\text{lin}} \cdot d \cdot 2 .
  \label{eq:naive-bpt}
\end{equation}
This cost does not depend on any interval. Hidden states stay in BF16 even when
the weights are quantized, as in Qwen3.6-27B-FP8.

\vspace{-0.05in}

\paragraph{\sys{}.}
\sys{} stores the input hidden state only at the entry of each group of
linear-attention layers, $A$ layers in total, and retains a fraction $\rho$ of
token positions:
\begin{equation}
  B_{\text{ours}} = A \cdot d \cdot 2\rho .
  \label{eq:ours-bpt}
\end{equation}
A model with $G$ full-attention layers has $G$ groups and $A = G - 1$: the
input of the first group is the token embedding, which needs no storage, and no
linear-attention layer follows the last full-attention layer. This gives $A=7$
for OLMo-7B and Qwen3.5-4B and $A=15$ for Qwen3.6-27B. With $\rho=1/16$,
$B_{\text{ours}}$ is 3{,}360, 2{,}240 and 9{,}600 bytes per token respectively,
a reduction of $(L_{\text{lin}}/A) / \rho
\approx 51$--$55\times$ over naive replay.

\begin{table}[!b]
  \centering
  \small
  \caption{Model parameters and the resulting sizes. All three models use
  convolution kernel width $w = 4$.}
  \label{tab:storage-params}
  \setlength{\tabcolsep}{4pt}
  \begin{tabular}{lrrr}
    \toprule
     & OLMo-7B & Qwen3.5-4B & Qwen3.6-27B \\
    \midrule
    $L_{\text{lin}}$            & 24 & 24 & 48 \\
    $d$                         & 3840 & 2560 & 5120 \\
    $H_k$ / $H_v$               & 30 / 30 & 16 / 32 & 16 / 48 \\
    $d_k$ / $d_v$               & 96 / 192 & 128 / 128 & 128 / 128 \\
    $C$                         & 11{,}520 & 8{,}192 & 10{,}240 \\
    \midrule
    $B_{\text{rec}}$            & 2{,}211{,}840 & 2{,}097{,}152 & 3{,}145{,}728 \\
    $B_{\text{conv}}$           & 69{,}120 & 49{,}152 & 61{,}440 \\
    $B_{\text{ckpt}}$           & 54{,}743{,}040 & 51{,}511{,}296 & 153{,}944{,}064 \\
    \midrule
    $B_{\text{native}}(8192)$   & 6{,}682.5 & 6{,}288 & 18{,}792 \\
    $B_{\text{naive}}$          & 184{,}320 & 122{,}880 & 491{,}520 \\
    Ratio                       & 27.6$\times$ & 19.5$\times$ & 26.2$\times$ \\
    \bottomrule
  \end{tabular}
\end{table}

Table~\ref{tab:storage-params} lists the parameters and the intermediate sizes.
For the two Qwen models, $B_{\text{ckpt}}$ equals the per-slot size of the
state pool that SGLang allocates at start-up, which we read from its logs. OLMo
is not supported by SGLang; we apply the same layout to the state shapes of its
reference implementation.

\end{document}